\documentclass[conference]{IEEEtran}
\IEEEoverridecommandlockouts
\usepackage{cite}
\usepackage{amsmath,amssymb,amsfonts}
\usepackage{graphicx}
\usepackage{textcomp}
\usepackage{xcolor}
\usepackage{epsfig}
\usepackage{amsmath}
\usepackage{amssymb}
\usepackage{color}
\usepackage{url}
\usepackage{multirow}
\newlength{\figurewidth}
\newlength{\smallfigurewidth}
\newcommand{\eatme}[1]{ }
\usepackage{paralist}
\usepackage{adjustbox}
\usepackage{wrapfig}

\usepackage{cite}
\usepackage{amsmath,amssymb,amsfonts}
\usepackage{algorithm}
\usepackage{graphicx}
\usepackage{textcomp}
\usepackage{xcolor}
\usepackage{subcaption}
\usepackage{siunitx}

\usepackage[T1]{fontenc}
\usepackage{graphicx}
\usepackage{booktabs}
\usepackage{longtable}
\usepackage{pbox}
\usepackage{mathtools}
\usepackage{caption}
\usepackage{subcaption}
\usepackage{algpseudocode}
\usepackage{algorithm}
\usepackage{siunitx}
\usepackage[inline]{enumitem}
\usepackage{siunitx}
\usepackage{diagbox}
\usepackage[]{mdframed}

\usepackage{comment}
\usepackage[normalem]{ulem}

\usepackage{xparse}
\NewDocumentCommand\N{sm}{\mathcal{N}\IfBooleanT#1{^{\ast}}_{#2}}

\makeatletter
\NewDocumentCommand{\@Coefficients}{m}{\text{\ttfamily\upshape #1}}
\newcommand\uMultilevelCoefficients{\@Coefficients{u\char`_mc}}
\newcommand\newMultilevelCoefficients{\@Coefficients{\~{u}\char`_mc}}
\makeatother

\begin{document}

\title
{\large
\textbf{Machine Learning Techniques for Data Reduction of CFD Applications 
}
} 

\author{Jaemoon Lee$^{\ast}$, Ki Sung Jung$^{\ddagger}$,  Qian Gong$^{\dag}$, Xiao Li$^{\ast}$, Scott Klasky$^{\dag}$, Jacqueline Chen${^\ddagger}$, Anand Rangarajan$^{\ast}$,\\ and Sanjay Ranka$^{\ast}$\\

\\[0.5em]
{\small\begin{minipage}{\linewidth}\begin{center}
\begin{tabular}{ccccc}
$^{\ast}$University of Florida & \hspace*{0.2in} & $^{\dag}$Oak Ridge National Laboratory &\hspace*{0.3in}${^\ddagger}$Sandia National Laboratories

\end{tabular}
\end{center}\end{minipage}}
}

\maketitle
\thispagestyle{empty}

\begin{abstract}
We present an approach called guaranteed block autoencoder that leverages Tensor Correlations (GBATC) for reducing the spatiotemporal data generated by computational fluid dynamics (CFD) and other scientific applications.  It uses a multidimensional block of tensors (spanning in space and time) for both input and output, capturing the spatiotemporal and interspecies relationship within a tensor. The tensor consists of species that represent different elements in a CFD simulation. To guarantee the error bound of the
reconstructed data, principal component analysis (PCA) is
applied to the residual between the original and reconstructed
data. This yields a basis matrix, which is then used to project
the residual of each instance. The resulting coefficients are
retained to enable accurate reconstruction. 

Experimental results demonstrate that our approach can deliver two orders of magnitude in reduction while still keeping the errors of primary data under scientifically acceptable bounds.
Compared to reduction-based approaches based on SZ, our method achieves a substantially higher compression ratio for a given error bound or a better error for a given compression ratio.
\end{abstract}

\section{INTRODUCTION}
Computational fluid dynamics (CFD) encompasses a broad range of  tools that are crucial for studying natural and engineered physical phenomenon that are critical to the nation’s energy security and economic competitiveness.  Direct numerical simulation (DNS) of turbulent combustion \cite{Chen2011,Vervisch2011, Driscoll2020,Apsden2011, Wen2023}, wherein all turbulence scales are numerically resolved, models blends of hydrogen with natural gas and ammonia to be used in power generation \cite{rieth2023direct, rieth22, Wiseman2021}, and DNS of hypersonic flows is being used to predict aerodynamics around complex adaptive geometries in environments with strong thermal and chemical nonequilibrium \cite{desai2022effects, DESAI2023112681, Martin2011}.

CFD simulation at the exascale Department of Energy (DOE) Leadership Class supercomputers runs on thousands of computational nodes powered by GPUs and generates massive volumes of data that must be stored and submitted to quanitity-of-interest (QoI) analysis. It is infeasible to store data at desired frequencies to capture the highly intermittent phenomena that occur in these transient simulations.

It is critical that trustworthy data reduction techniques be developed for reducing data generated in areas such as aerospace engine design, hypersonics, dispatchable power generation with hydrogen blends, and turbulent and reactive flows. Trustworthiness necessitates that the compressors provide guarantees on reduction-incurred errors. CFD with multiphysics also suffers from the curse of dimensionality, e.g., transporting hundreds of species. The multiphysics is also highly nonlinear, for instance, this occurs during autoignition processes in combustion applications where exponential growth of species concentrations depends upon the local strain rate and mixing conditions. Thus, reduction techniques have to take this into account.

The focus of this paper is to demonstrate that ML-based autoencoders offer the following benefits:

\begin{compactenum}

\item	{\it Leverage Spatiotemporal Relationships in Mesh Structures:}  CFD  datasets have underlying data structures that consist of structured and block-structured multidimensional meshes. It is important that the data reduction techniques address strong spatiotemporal correlations that are naturally present in these structures. Notably, the values corresponding to each species may increase or decrease exponentially over the course of the simulation.
\item {\it Leverage Relationships within Tensors Representing Physics at Individual Grid Points:} CFD simulations update and store tensors comprising several tens to a hundred species and their attributes at each grid point or particle. Reduction techniques should be able to leverage relations of elements within a tensor. 

\end{compactenum}

The proposed guaranteed block autoencoder (GBATC)  utilizes a multidimensional block of tensors in space and time to capture the spatiotemporal relationships as well as the interspecies relations within a tensor. 
The original data are divided into smaller spatiotemporal blocks. We employ a 3D convolutional autoencoder to capture the spatiotemporal correlations within each block. To improve the
compression quality further, we introduce a tensor correction
network. After training the AE, we
convert each reconstructed instance by the AE, comprising S
number of 3D blocks, into a set of S-dimensional tensors.
The tensor correction network takes the reconstructed tensors
and learns a reverse point-wise (in temporal and spatial space)
mapping from the reconstructed tensors to the original tensors.

To guarantee the error bound of the reconstructed data, principal component analysis (PCA) is applied to the residual between the original and reconstructed data. This yields a basis matrix, which is then used to project the residual at each instance. The resulting coefficients are retained to ensure an accurate recovery of the residual values. The number of coefficients saved is incrementally increased until the error bound is satisfied. Additionally, quantization and entropy coding techniques are applied to both the latent data from the GAE and the PCA coefficients. This further improves the compression ratio of the overall process.

The data reduction method is validated using simulation data generated by Sandia's compressible reacting direct numerical simulation (DNS) code, S3D \cite{Chen09}. The conservation equations for continuity, momentum, total energy, and species continuity describing reacting flows are solved using an 8th-order spatial finite difference method and 4th-order explicit time integrator in S3D. Note that these chemically reacting flow simulations have high dimensionality, involving detailed chemical mechanisms that contain a large numbers of species ($\sim O(10^3)$) and elementary reactions ($ \sim O(10^4)$), requiring a vast number of species transport equations that need to be solved. However, correlations exist between species in turbulent flames and, hence,  reduced-order modeling (mostly using linear PCA) is increasingly being adopted in combustion simulations to reduce the dimensionality of the composition space \cite{Cuoci21,Jung24,Savarese24,Kumar23}. Hence, the overall efficiency of data reduction can be significantly enhanced by exploiting spatiotemporal correlations among species.

Experimental results demonstrate that our approach can derive two to three orders of magnitude in reduction while still maintaining the errors of primary data under scientifically acceptable bounds.
In comparison to previous research \cite{SZ_3}, our method achieves a substantially higher compression ratio.  CFD Scientists are also  interested in quantities of interest (QoIs) that are derived from raw data. Thus, it is important that the reduction methods provide reasonable error bounds on them. One of the crucial QoIs used in the species transport equation is the net production rate for each species (which involves reactions with other species) with the rate being dependent on the forward and reverse rate constants of the reactions underlying the net production. The forward and reverse reaction rate constants are pointwise estimations and follow an Arrhenius equation, which is a nonlinear function of local temperature, pressure, and concentrations of the species. We show that our approach delivers smaller errors on these QoIs than SZ under the same compression ratios.

The remainder of the paper is organized as follows. Section ~\ref{sec:methodology} describes our compression pipeline. Experimental results using our method are presented in Section~\ref{sec:experiment}  for different levels of compression and accuracy. Section~\ref{sec:relwork} presents related work. Conclusions are provided in Section~\ref{sec:conclusion}.

\section{Methodology}\label{sec:methodology}

\subsection{Guaranteed Autoencoder (GAE)} 
An autoencoder (AE) is a neural network designed to learn efficient representations of data by compressing the input into a lower-dimensional latent space and then reconstructing the original input from this compressed representation. It consists of two main parts: an encoder and a decoder. The encoder compresses the input data from high-dimensional to lower-dimensional latent space representation, while the decoder reconstructs the original input from this compressed representation. \cite{jaemoon}

AEs can be trained for data reduction by minimizing the reconstruction errors between original and reconstructed data. In this study, we employ the standard mean-squared error (MSE) loss function to measure compression-incurred errors in PD. 

After training, the decoder and the latent representations  $\boldsymbol{H} = \left\{\boldsymbol{h}_{i}\right\}_{i=1}^{N}$ need to be stored. We store the decoder without compression, given its small size.  However, storing floating-point latent vectors is not an efficient approach. To overcome this challenge, we employ a compression technique that involves float-point quantization followed by entropy encoding to compress the latent space data. Our approach to enhancing compression efficiency employs a two-step strategy. Firstly, we uniformly quantize these coefficients into discrete bins, each with a bin size of $d$. This discretization process effectively represents all values within each bin by its central value, transforming the originally continuous data into a discrete form. Subsequently, we utilize Huffman coding to compress these quantized coefficients. Huffman coding assigns shorter codes to frequently occurring quantized coefficients, optimizing the representation of the data and achieving higher compression efficiency. This combined technique significantly reduces the data's size while retaining crucial information.

\begin{figure}
  \centering
    \includegraphics[width=\columnwidth]{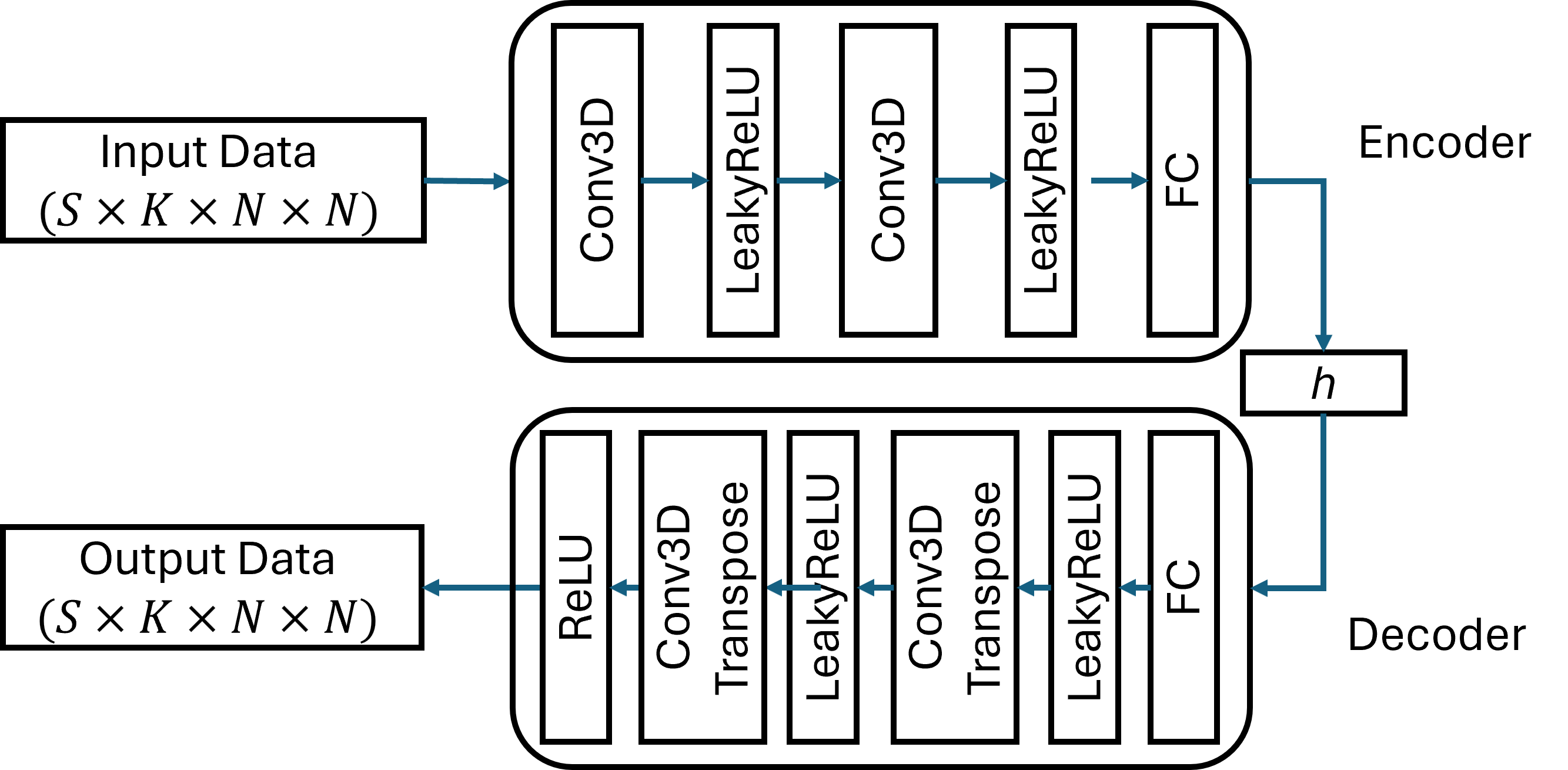}
    \caption{The structure of the autoencoder: Conv3D denotes the 3D convolution layer, Con3DTranspose denotes the 3D transposed convolution layer, FC denotes the fully connected layer, and $h$ denotes the latent space. Leaky ReLU is adopted as the activation function. Each channel in 3D convolution layers processes each species of $S$ species data in the CFD application.}
    \label{fig:ae}
     \vspace*{-0.5cm}
\end{figure}

\paragraph*{Bounding the Reconstruction Error}

We aim to limit reconstruction errors for all instances in an AE. Although any appropriate error bound can be applied within our framework, our emphasis lies in constraining the $\ell_{2}$ norm of each instance residual, denoted as $\left\|\boldsymbol{x}-\boldsymbol{x}^{R}\right \|_{2}$. To optimize compression ratios, we only bound the AE reconstruction error for instances whose residual $\ell{2}$ norm exceeds the specified threshold $\tau$.
After obtaining the reconstructed data from the autoencoder, we apply PCA to the residual of the entire dataset to extract the principal components or basis matrix, denoted as 
${U}$. These basis vectors are sorted in descending order according to their corresponding eigenvalues. These principal components capture the directions of maximum variance in the residual data. Although we compress the data block by block, we treat each patch of data as a single instance and compute the basis matrix at the patch level. To guarantee the error bound for each patch of data, we project the residual of each patch of data onto the space spanned by $U$ and select the leading coefficients such that the $\ell_2$ norm of the corrected residual falls below the specified threshold $\tau$. These coefficients, representing the residual, are derived from the equation: 
\begin{equation}
\boldsymbol{c}=U^{T}\left(\boldsymbol{x}-\boldsymbol{x}^{R}\right).
\end{equation}
It's important to note that the complete recovery of the residual $\boldsymbol{x}-\boldsymbol{x}^{R}$ can be achieved by computing $\boldsymbol{Uc}$, yielding the coefficient vector $\boldsymbol{c}\equiv \left[c_{1},\ldots,c_{D}\right]$.  Given that the error bound criterion is based on $\ell_{2}$, we compute $\{c_{k}^{2}\}_{k=1}^{D}$ and sort the positive values, resulting in coefficients ordered from the largest to the least contribution to errors. The top $M$ coefficients and corresponding basis vectors are selected to satisfy the target error bound $\tau$. To minimize the storage cost of these coefficients, we compress the selected coefficients $\boldsymbol{c}_{s}$ using a method similar to that employed for compressing AE latent coefficients. These coefficients are quantized before being used for reconstructing the residual. The corrected reconstruction $x^{G}$ is
\begin{equation}
\boldsymbol{x}^{G}=\boldsymbol{x}^{R}+U_{s}\boldsymbol{c}_{q},
\end{equation}
where $\boldsymbol{c}_{q}$ is the set of selected coefficients $\boldsymbol{c}_{s}$ after quantization and $U_{s}$ is the set of selected basis vectors. We increase the number of coefficients until we achieve $\left\|\boldsymbol{x}-\boldsymbol{x}^{G}\right\|_{2}\leq \tau$. Therefore, in order to guarantee the error bound, we need to store the basis matrix, the selected coefficients $\boldsymbol{c}_{q}$ for each patch of data and their basis vector indices.

\subsection{Guaranteed Block Autoencoder (GBA)}
To explore the spatiotemporal correlation within scientific data, we integrate 3D convolution into our AE architecture that incorporates both temporal and spatial correlations.
By harnessing the capabilities of 3D convolutional operations, our autoencoder excels in capturing intricate spatial patterns and temporal dynamics simultaneously. In the CFD application, there are multiple species that have different chemical nature. For each species, we partition the original data into non-overlapping $ N \times N$ patches at each data frame. Then, we group $K$ consecutive patches from the same location across time into a single block of data as an input to the AE. Each instance that is processed by the AE consists of a set of blocks that lie in the same temporal and spatial space across all the species. Because of the different nature of species, each channel of 3D convolution layers processes each species data, and all the features across all the species obtained by convolutional layers are compressed together with a fully connected layer. The structure of the AE is shown in Figure \ref{fig:ae}.

\begin{figure}
  \centering
    \includegraphics[width=0.9\linewidth]{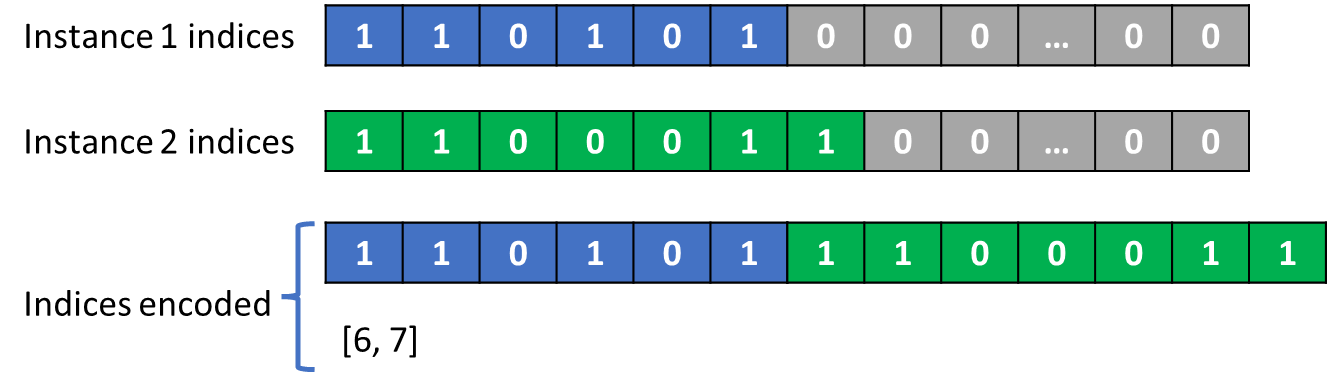}
    \caption{Indices encoding}
    \label{fig:indices}
     \vspace*{-0.5cm}
\end{figure}

In the GBA, the reconstruction errors are guaranteed based on the block. We follow the same GAE post-processing algorithm described in \ref{alg:GAE}, but here each data instance becomes a block that is converted into a vector. Also, we apply the post-processing to each species separately. To store the basis vector indices for each patch of data effectively, we start by encoding the indices using a binary sequence. Remarkably, we notice that the initial basis vectors (associated with larger eigenvalues) are selected more frequently. As a result, these binary sequences frequently end with a sequence of zeros. We decide to store only the shortest prefix subsequence that contains all ones. Furthermore, we introduce an additional value to indicate the length of this subsequence. An illustrative example is provided in Figure \ref{fig:indices}.

\subsection{GBA with Tensor Correction Network (GBATC)}
\begin{figure*}[ht!]
  \centerline{\includegraphics[width=160mm]{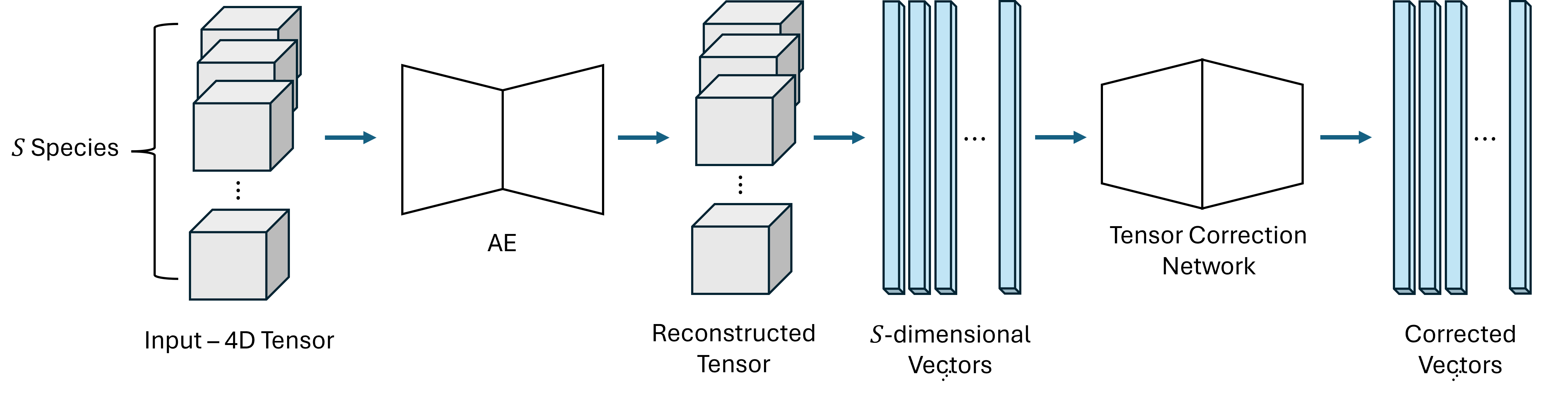}}
  \caption{Guaranteed Block Autoencoder with Tensor Correction Network (GBATC). The AE processes 3D blocks through convolutional layers with $S$ channels and further compresses the block with a fully connected layer as described in Figure~\ref{fig:ae}. After getting the reconstructed data, we convert the block into a set of vectors. The vectors represent $S$ species data for the specific temporal and spatial points and they are corrected by the tensor correction network. The network learns a mapping from the reconstructed data back to the original data, and it is overcomplete as compression is performed by the AE.}\label{GBATC_arch}
\end{figure*}

In the GBA, we utilize a set of blocks as temporal and spatial correlation are effectively addressed by convolutional layers followed by a fully connected layer. This enables us to achieve a high data reduction amount. To improve the compression quality further, we introduce a tensor correction network as described in Figure~\ref{GBATC_arch}. After training the AE, we convert each reconstructed instance by the AE, comprising $S$ number of 3D blocks, into a set of $S$-dimensional tensors. The tensor correction network takes the reconstructed tensors and learns a reverse point-wise (in temporal and spatial space) mapping from the reconstructed tensors to the original tensors. We employ an overcomplete network, where the hidden layer sizes are not smaller than the input tensor size. This is because the dataset is already compressed by the AE, and the tensor correction network adjusts the reconstructed data from the AE to become closer to the original dataset. The advantage of incorporating the tensor correction network is that there are no additional hidden representations to be stored. We only need to store the network parameters. Once we get the adjusted dataset, we apply the same block-based post-processing approach used in the GBA.

\begin{algorithm}
\caption{The GAE Algorithm}
\label{alg:GAE}
\begin{flushleft}
\textbf{Input:} Input data $\boldsymbol{X} = \left\{\boldsymbol{x}_{i}\right\}_{i=1}^{N}$, reconstructed data $\boldsymbol{X}^R = \left\{\boldsymbol{x}_{i}^{R}\right\}_{i=1}^{N}$, target error bounds $\left\{\tau_{i}\right\}_{i=1}^{N}$.

\textbf{Output:} Corrected reconstruction $\boldsymbol{X}^G = \left\{\boldsymbol{x}^{G}_{i}\right\}_{i=1}^{N}$, coefficients $\mathbf{C} = \left\{\boldsymbol{c}_{i}\right\}_{i=1}^{N}$, indices $\mathbf{I} = \left\{I_{i}\right\}_{i=1}^{N}$ where $I_i$ is an index set, basis matrix $U$.
\end{flushleft}
\begin{algorithmic}[1]
\State Run PCA on the residual $\boldsymbol{X}-\boldsymbol{X}^R$, obtaining basis matrix $U$
\For{$i=1$ \textbf{to} $N$}  
\State $\boldsymbol{x}\gets \boldsymbol{x}_{i}$, $\boldsymbol{x}^{R}\gets \boldsymbol{x}_{i}^{R}, \tau \gets \tau_i$.
\State Compute $\ell_{2}$ norm $\delta=\left\|\boldsymbol{x}-\boldsymbol{x}^{R}\right\|_{2}$.
\If{$\delta > \tau$}
\State Project residual $\boldsymbol{c}=U^{T}(\boldsymbol{x}-\boldsymbol{x}^{R})$ and sort $c_{k}^{2},~\forall k$.
\State $M\gets 1$
\While{$\delta>\tau$}
\State $\boldsymbol{c}_{s},U_{s}\gets$ Top $M$ coefficients in $\boldsymbol{c}$ and corresponding basis vectors in $U$.
\State $\boldsymbol{c}_{q}$ $\gets$ Quantize($\boldsymbol{c}_{s}$)
\State $\boldsymbol{x}^{G}\gets \boldsymbol{x}^{R}+U_{s}\boldsymbol{c}_{q}$.
\State $\delta \gets \left\|\boldsymbol{x}-\boldsymbol{x}^G\right\|_{2}$
\State $M\gets M+1$
\EndWhile
\State $\boldsymbol{c}_{i}\gets \boldsymbol{c}_{q}$
\State $I_{i}\gets$ Index set for $\boldsymbol{c}_{q}$
\State $\boldsymbol{x}^{G}_{i}\gets \boldsymbol{x}^{G}$
\EndIf
\EndFor
\end{algorithmic}
\end{algorithm}

\subsection{SZ}

SZ is a prediction-based compression model where each data point is predicted based on its adjacent data points. It consists of four stages: prediction, quantization, Huffman coding, and lossless compression. Once a data point is predicted, the difference between the original value and the predicted value is quantized in a linear scale according to user-specified error bounds. Then, the quantization array is stored in a lossless manner using Huffman encoding. The adjacent data points must be decompressed data to ensure the error bounds at the decompression stage, which means the prediction process is finished for those adjacent data points. Various versions of SZ have been developed using different prediction methods to improve compression quality. SZ1.4 \cite{SZ1.4} uses a Lorenzo predictor that predicts a data point using a linear combination of its adjacent data points. In SZ2 \cite{SZ2.0}, either a Lorenzo predictor or a linear regression predictor is selected according to the prediction accuracy. For the regression predictor, the entire dataset is split into $6 \times 6 \times 6$ for a 3D dataset or $12 \times 12$ for a 2D dataset, and then coefficients of a linear regression model are computed for each data block. In \cite{SZspline}, dynamic spline interpolation is developed for prediction. From linear to cubic spline interpolation is selected according to the prediction accuracy that is usually affected by the data dimensionality and error bounds. SZ3 \cite{SZ3} leverages all the prediction methods and chooses one of them based on accuracy.

\section{Experimental Results}\label{sec:experiment}

This section presents the experimental results of applying our compression pipeline to the dataset obtained by the S3D application. We detail the evaluation metrics and describe key characteristics of the data set, including PD and QoI. We establish a baseline method for comparison with our approach. We then compare the accuracy and effectiveness of our compression method with the SZ (SZ3), one of the leading state-of-the-art lossy compression techniques for scientific data compression.

\paragraph*{Evaluation Metrics}

We utilize a relative error criterion, the normalized root mean square error (NRMSE), to assess reconstruction quality because species lie in different data ranges. To evaluate overall compression quality, we measure NRMSE per species and take the average of NRMSEs of all the species. The NRMSE is defined as
\begin{equation}
\mathrm{NRMSE}\left(\boldsymbol{X},\boldsymbol{X}^{R}\right)=\frac{\sqrt{\sum_{i=0}^{N}\sum_{d=0}^{D}\left(x_{i,d}-x_{i,d}^{R}\right)^{2}/ND}}{\max\left(\boldsymbol{X}\right)-\min\left(\boldsymbol{X}\right)},
\end{equation}
where $\boldsymbol{X}$ and $\boldsymbol{X}^{R}$ denote the original and reconstructed data of a single species that has $N$ data instances, respectively. The compression ratio is defined as the ratio between the bytes of PD and compressed outputs. The compressed output comprises the encoded representation of the AE encoder, encoded coefficients with their corresponding basis indicators of the post-processing, network parameters, and all the dictionaries for entropy coding. Those outcomes are required to ensure that the errors in reconstructed PD adhere to user-prescribed error tolerance.

\paragraph*{S3D Dataset}


We briefly present results on the dataset that represents the compression ignition of large hydrocarbon fuels under conditions relevant to homogeneous charge compression ignition (HCCI), as introduced in \cite{Yoo11}.  In HCCI combustion, isentropic compression of the initial mixture in a constant volume results in sequential low-temperature autoignition followed by high-temperature autoignition.  Both stages are modulated by turbulent strain and mixing which produces pockets of temperature and composition inhomogeneities. The spatiotemporal mixture nonuniformities result in significant variances in ignition delay as monitored by species mass fractions and temperature.  The dataset comprises a two-dimensional space of size 640$\times$640, collecting data over 50 time steps uniformly from $t =$ 1.5 to 2.0 ms, where intermediate-temperature chemistry is clearly observed. A 58-species reduced chemical mechanism \cite{Yoo11} is used to predict the ignition of a fuel-lean $n$-heptane+air mixture. Thus, each tensor corresponds to 58 species. One of the crucial QoIs for the species transport equation is the production rate for each species (and this involves other species as well) with the rate being dependent on the forward and reverse rate constants of the reactions underlying the production. The forward and reverse reaction rate constants are pointwise estimations and follow an Arrhenius equation, which is a nonlinear function of local temperature, pressure, and species concentrations. Therefore, the QoIs are $\mathcal{O}(N)$. The spatiotemporal blocking was chosen to capture both the spatiotemporal relationships and relationships between the species. The data were divided into spatiotemporal blocks consisting of four time steps and $5\times 4$ location blocks for all 58 species. Concomitant QoIs, production rates for each species, are available via computation using the open-source Cantera library \cite{cantera}.

\begin{figure*}
\begin{subfigure}[t]{1.0\linewidth}
    \centering
    \includegraphics[width=0.4\linewidth]{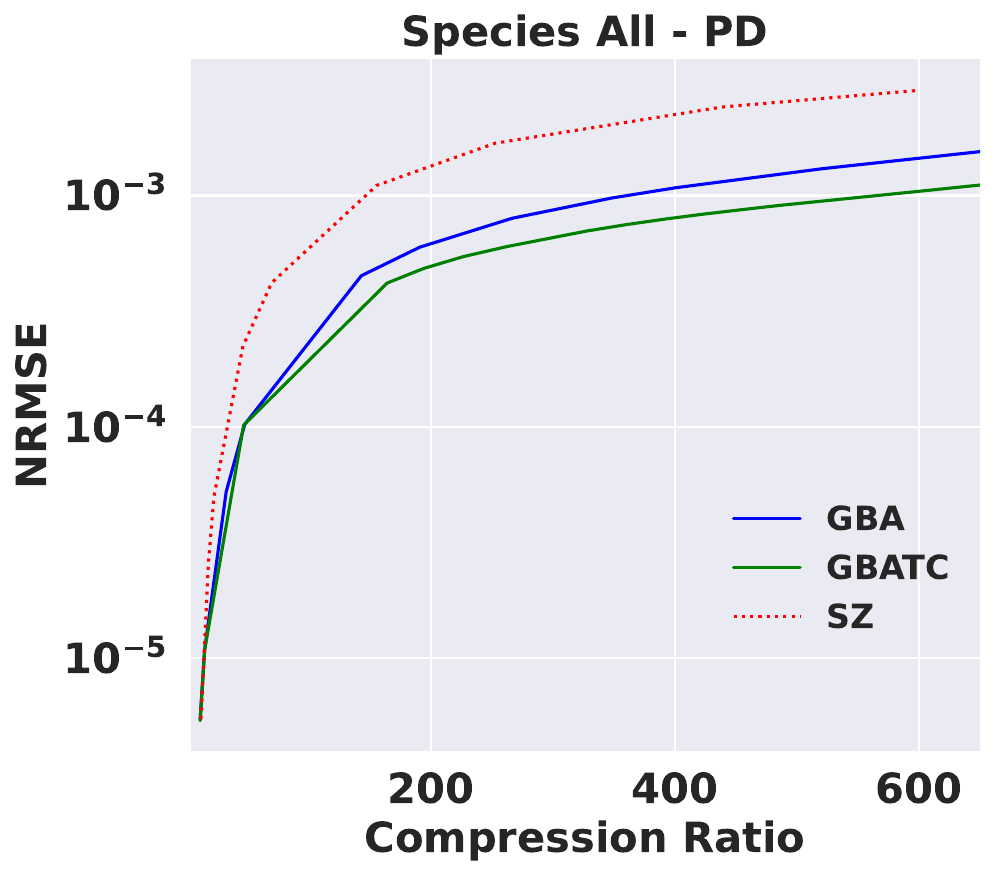}
    \hspace{0.5cm}
    \includegraphics[width=0.4\linewidth]{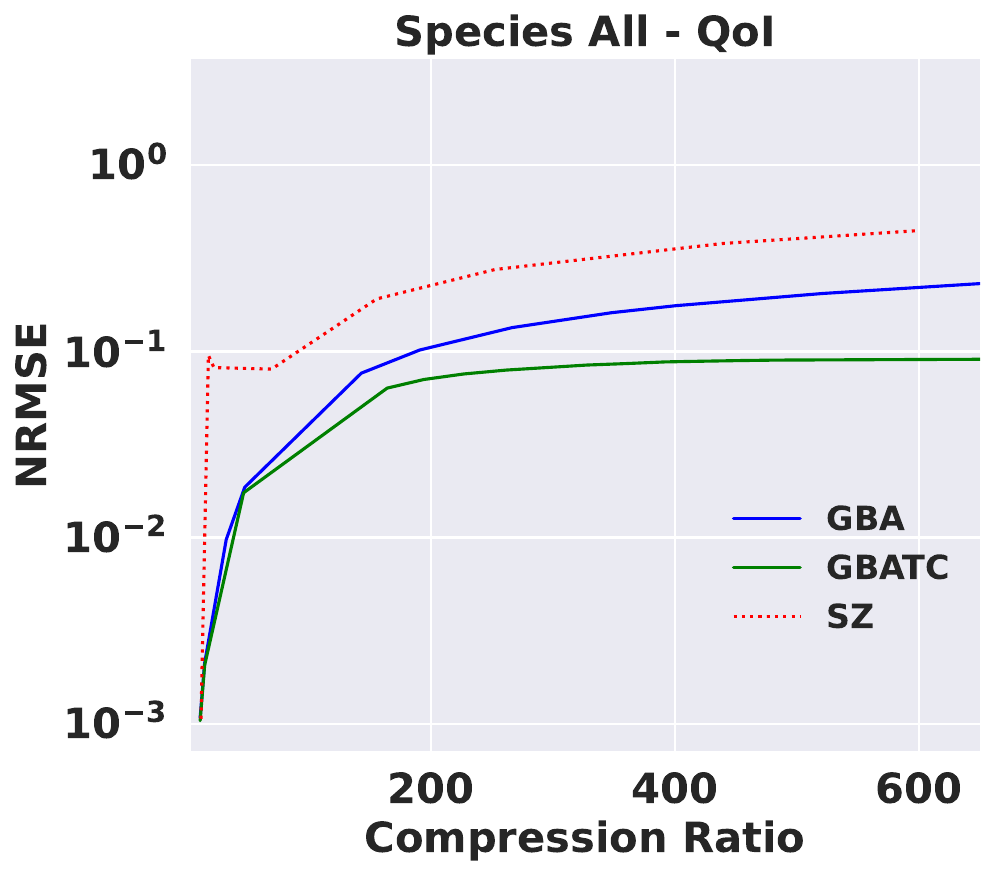}
    \vspace{-0.2cm}
  \end{subfigure}
  \caption{Comparison of a block-based GAE with SZ when the QoIs are $\mathcal{O}(N)$ with only PD guarantees: (a) PD error versus compression ratio, (b) QoI error versus compression ratio. Our approach has high compression ratios because it utilizes the entire tensor along with spatiotemporal relationships.}
    \label{fig:gaes3d}
    \vspace{-.1cm}
\end{figure*}

\paragraph*{Results}
We compress the mass fraction data (denoted as PD) across 58 species utilizing the GBA, GBATC, and SZ. In both the GBA and GBATC, the convolutional AE in Figure~\ref{fig:ae} is trained with $58\times 5\times 4\times 4$ blocks of tensors. The 58 species are treated as individual channels for the 3D convolutional layers in the AE. For each channel, $5\times 4\times 4$ blocks representing temporal and 2D spatial dimensions are processed through 3D convolutional layers in the AE encoder. All the features obtained through convolutional layers are compressed together using a single fully connected (fc) layer. We opt for one fc layer in the AE encoder as additional fc layers do not enhance compression accuracy for this application. The latent size of the AE encoder is set to 36. The compressed representations are then quantized followed by Huffman coding as described in Section~\ref{sec:methodology}. Subsequently, the AE decoder reconstructs the data from these encoded representations. The GBATC has an additional tensor correction network. Instead of using block scheme, the network learns a point-wise mapping from the 58 reconstructed species to the 58 original species. The tensor correction network consists of four fc layers. The fc layers convert 58 units to 232 units, 232 units to 464 units, 464 units to 232 units, and 232 units to 58 units with Leaky ReLU activations. The network is employed to adjust the reconstructed data from the AE to improve the overall NRMSE error. 

Then, we apply the block-based post-processing on the reconstructed data by the AE for both GBA and GBATC to guarantee reconstruction errors using the approach presented in Section ~\ref{sec:experiment}. During the post-processing, each species is adjusted individually using a PCA basis derived from all the $5\times 4\times 4$ residual blocks for that species, resulting in a $80\times 80$ dimensional PCA basis stored per species.

Figure~\ref{fig:gaes3d} illustrates the evaluation of the compression results produced by the GBA, GBATC, and SZ (note that the NRMSE results are plotted on a log scale). The reaction rates of the 58 species, denoted as QoI, are computed using the Cantera library with the reconstructed PD from the GBA, GBATC, and SZ. The results show that both the GBA and the GBATC outperform the SZ. Both these approaches achieve a greater compression ratio at equivalent levels of reconstruction errors. Notably, at a PD NRMSE of $10^{-3}$, which is the accuracy recommended by domain experts, the GBA and GBATC achieve compression ratios of 400 and 600 respectively, significantly higher than a compression ratio of 150 achieved by SZ. This can be partially attributed to  efficient nonlinear mapping of the PD tensor onto a smaller manifold. SZ  compresses each scalar separately. The AE can capture global correlations across tensor blocks, resulting in a higher degree of compression. The GBATC has better NRMSE error as compared to GBA for a given compression ratio. This shows that the tensor correction network learns an effective reverse mapping from 58 reconstructed species to 58 original species.

As for the QoI, the errors of the QoI derived from the reconstructed PD of the GBA and GBATC are also considerably superior to SZ. This is mainly attributable  to the lower PD errors. 
\eatme{
SZ cannot achieve any QoI NRMSE below $10^{-1}$ unless the NRMSE of the reconstructed PD is smaller than $10^{-5}$. This discrepancy between the accuracy of the reconstructed PD and its QoI in the SZ may lead to erroneous post-analysis for the S3D application. These findings, while promising, are still preliminary and need further investigation. \textcolor{blue}{Maybe here to insert a sentence, saying that we measured the averaged errors across 58 species. We wonder whether the bad QoI errors in SZ (shown in Figure 3) was caused by a few outlier species, so we conducted the following investigations...}
}

\begin{figure*}[ht!]
  \centerline{\includegraphics[width=160mm]{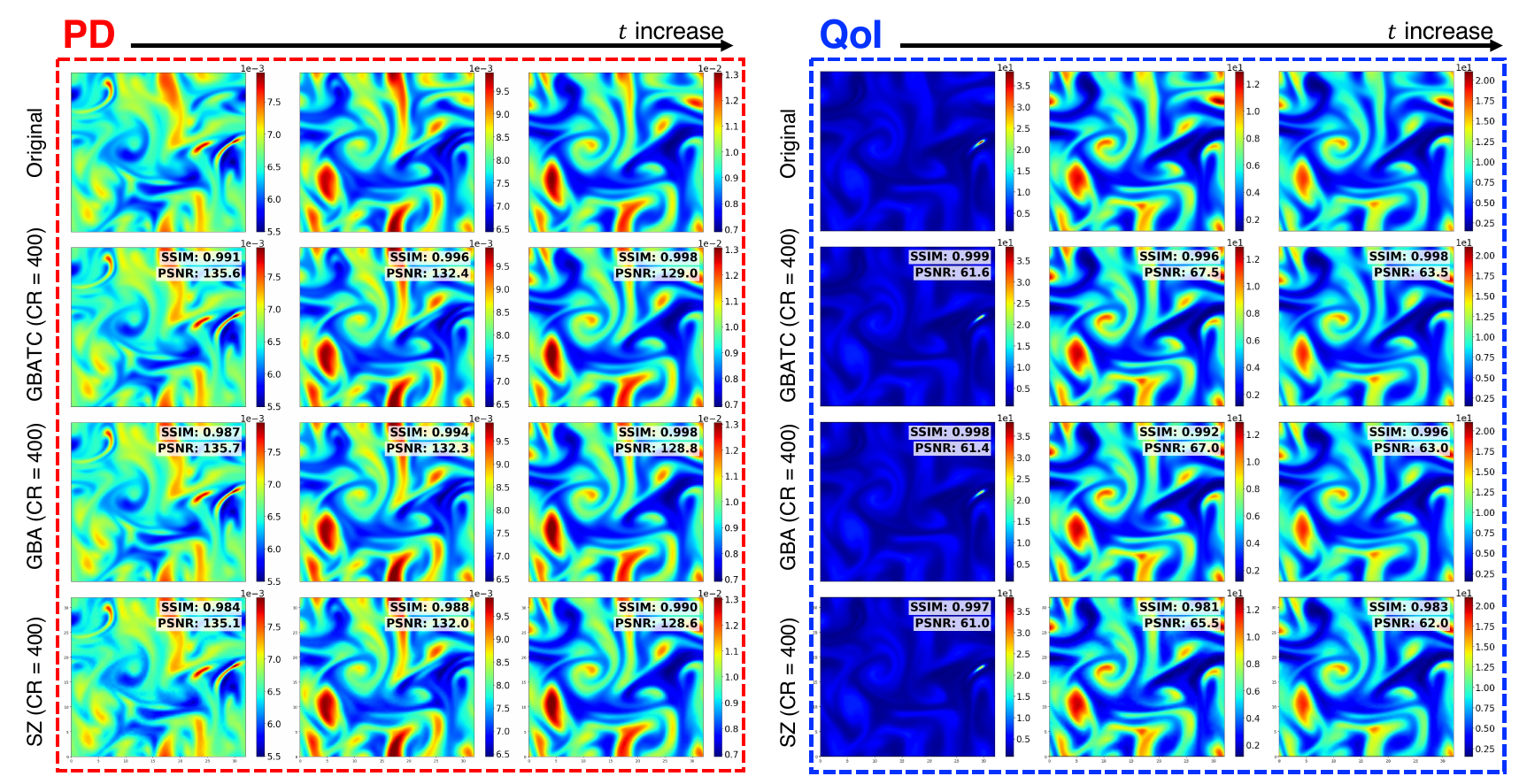}}
  \caption{Temporal evolution ($t$ = 1.5, 1.8, and 2.0 ms) of the (left half) mass fraction and (right half) formation rate of H$_2$O as predicted by (first row) DNS, (second row) GBATC, (third row) GBA, and (last row) SZ. The compression ratios for all the results are 400.}\label{H2O_summary}
\end{figure*}

We now investigate  the accuracy of data compression using three different methods, GBATC, GBA, and SZ at a species level.  Figure~\ref{H2O_summary} presents the temporal evolution of the mass fraction and formation rate of major species, represented by H$_{2}$O, as predicted by the original DNS and decompressed datasets using the aforementioned compression methods. Note that major species include the reactants and products (i.e., nC$_7$H$_{16}$, O$_2$, CO$_2$, CO, and H$_2$O), whereas minor species includes radical species (smaller concentrations) that lead to chain branching and ultimately ignition. The decompressed dataset is based on a compression ratio of 400. 

In the combustion community, numerous studies \cite{Jung24,Savarese24,Cuoci21} have demonstrated the accuracy of applying linear PCA to reduce the dimensionality of the composition space, when the target range for the NRMSE of the reconstructed dataset is between 0.03–-0.01 (i.e., the low-dimensional manifold retains 99.9–99.99\% of the total variance of the original dataset). Therefore, the targeted NRMSE in the present study, set at near to 0.001, is a reasonable value that ensures accuracy of the decompressed dataset. It is evident from Fig.~\ref{H2O_summary} that there is no discernible difference of PD between the original and decompressed dataset using GBATC, GBA, and SZ. The corresponding QoI, the formation rate of H$_2$O, looks to show good agreement between the original and decompressed datasets. 

To further quantify the actual error between the original and the decompressed dataset, we employ the structural similarity index measure (SSIM) and peak signal-to-noise ratio (PSNR). SSIM is a metric that quantifies the perceptual differences between two images \cite{Wang04}, and similarly, PSNR refers to the ratio of the maximum possible value of a signal to the value of the corrupting noise that impacts its quality of representation. Higher values of SSIM or PSNR indicate better quality of the reconstructed image. Our findings show that the accuracy of reconstructing H$_2$O and evaluating QoI is highest with the GBATC method, followed by GBA and SZ. Note that using linear PCA, the rank of the low-dimensional manifold, for which the NRMSE is 0.001, is 46 (where the rank of the full-order model is 58) for the given dataset. The high rank is an indication of the inherent complexity of the original dataset and highlights the efficiency of the data compression techniques described in this paper. 

\begin{figure*}[ht!]
  \centerline{\includegraphics[width=160mm]{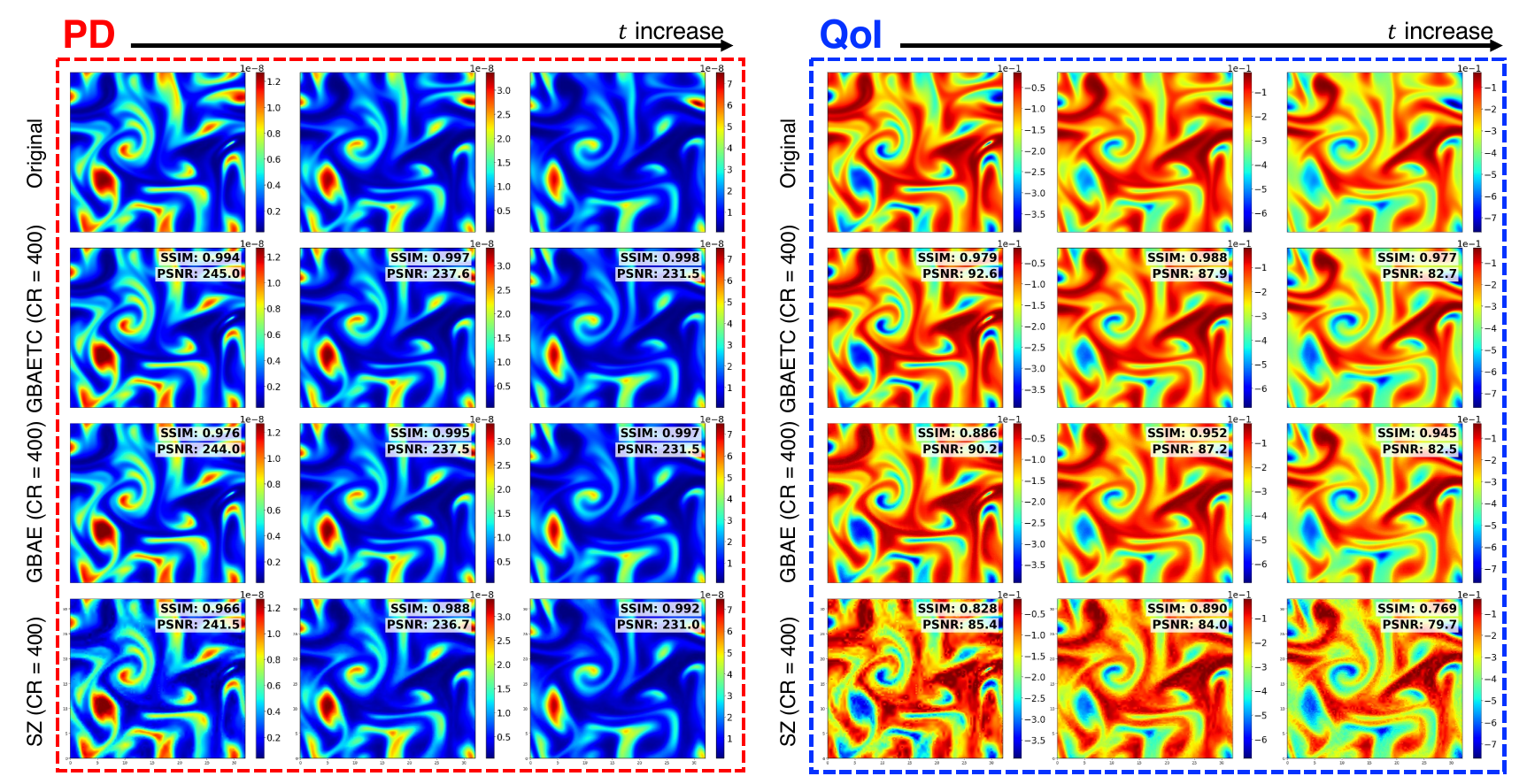}}
  \caption{Temporal evolution ($t$ = 1.5, 1.8, and 2.0 ms) of the (left half) mass fraction and (right half) formation rate of C$_2$H$_3$ as predicted by (first row) DNS, (second row) GBATC, (third row) GBA, and (last row) SZ. The compression ratios for all the results are 400.}\label{C2H3_summary}
\end{figure*}

Figure~\ref{C2H3_summary} compares the temporal evolution of the mass fraction and formation rate of the minor species, represented by C$_2$H$_3$, obtained from the original and decompressed datasets using different data compression methods with a compression ratio of 400. While GBATC and GBA show reasonable accuracy in reproducing both PD and the QoI for minor species, SZ exhibits a noticeable discrepancy in deriving QoI from the reconstructed data, illustrating superior performance of GBATC or GAE over SZ. Note that both SSIM and PSNR for GBATC are slightly higher than those for GBA both in PD and QoI. It is also noted that the overall error of QoI for minor species is higher than that for major species, as shown in Figs.~\ref{H2O_summary}--\ref{C2H3_summary}, mainly because minor species are more sensitive to errors in PD. 

\begin{figure*}[ht]
  \centerline{\includegraphics[width=0.75\linewidth]{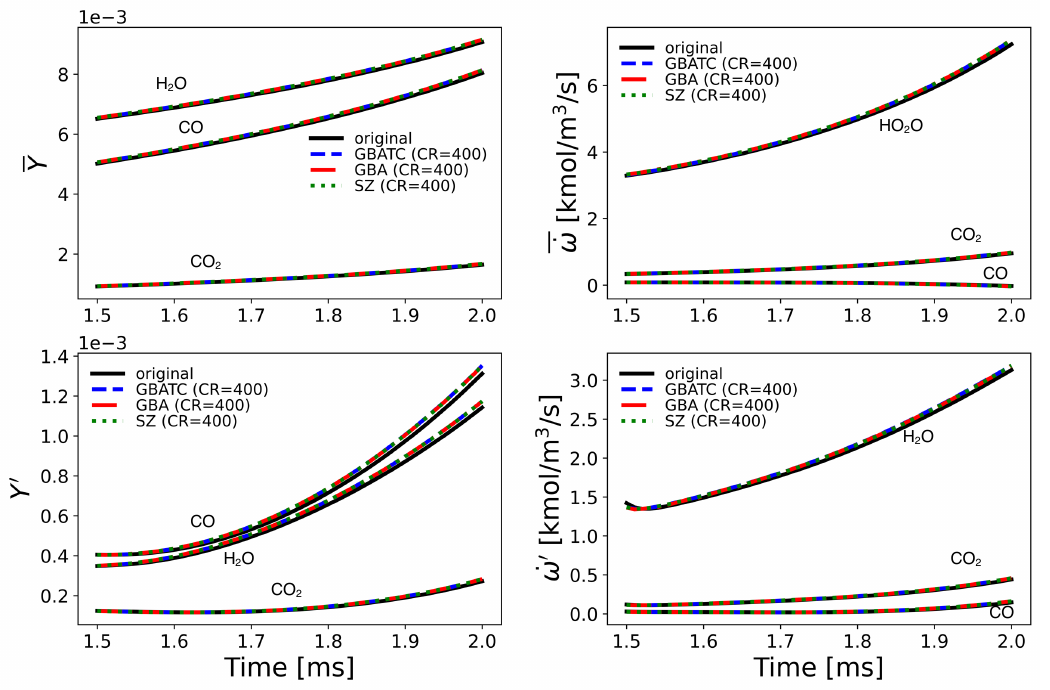}}
  \caption{Variations in the (top) mean and (bottom) standard deviation of the (left) mass fraction and (right) formation rates of the major species, represented by H$_2$O, CO, and CO$_2$, over time, as predicted by (solid) DNS, (dashed) GBATC, (dashed-dot) GBA, and (dotted) SZ. The compression ratios for all methods are 400.}\label{mean_std_major}
\end{figure*}

\begin{figure*}[ht]
  \centerline{\includegraphics[width=0.75\linewidth]{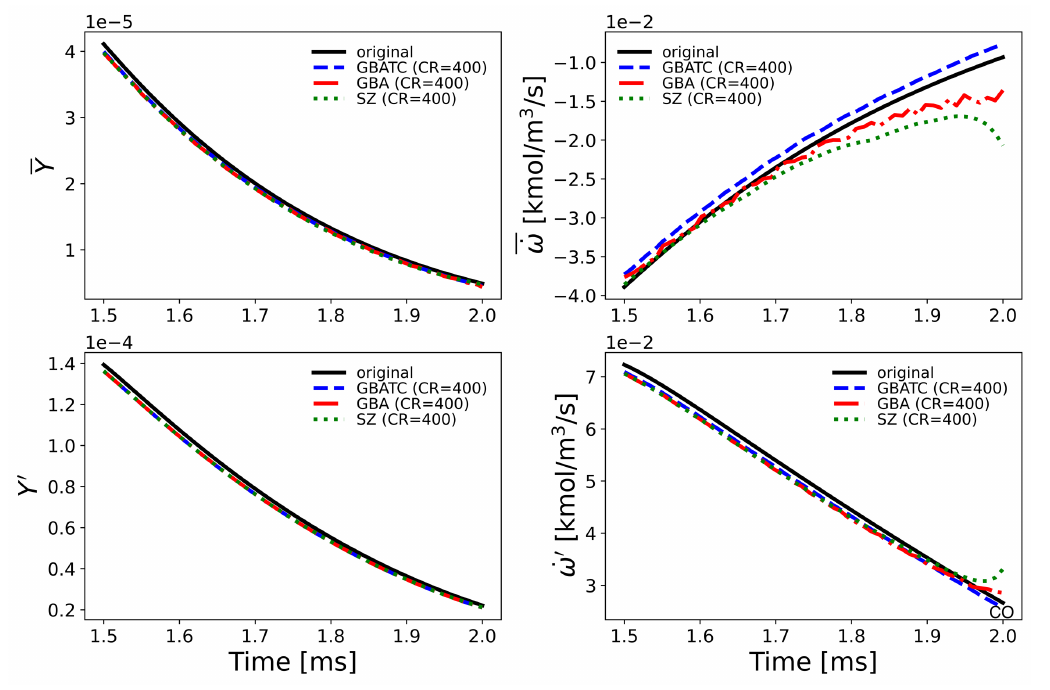}}
  \caption{Variations in the (top) mean and (bottom) standard deviation of the (left) mass fraction and (right) formation rates of the minor species, represented by nC$_3$H$_7$COCH$_2$, over time, as predicted by (solid) DNS, (dashed) GBATC, (dashed-dot) GBA, and (dotted) SZ. The compression ratios for all methods are 400.}\label{mean_std_minor}
\end{figure*}

Figure~\ref{mean_std_major} presents the variations in the mean and standard deviation of both mass fractions and formation rates of several major species, including CO, CO$_2$, and H$_2$O, predicted by DNS and the decompressed dataset using three different data compression methods with a compression ratio of 400. Consistent with the results in Fig.~\ref{H2O_summary}, the original and decompressed datasets are in good agreement in predicting major species, irrespective of the data compression methods. Figure~\ref{mean_std_minor} also illustrates the variations in the mean and standard deviation of another minor species, nC$_3$H$_7$COCH$_2$, predicted by DNS and different data compression methods. Note that nC$_3$H$_7$COCH$_2$ is a representative low-temperature ignition species and its mass fraction is very small during the period of high-temperature ignition, $t = 1.5-2.0$ ms. 

Fig.~\ref{mean_std_minor} highlights the differences between GBATC and the other methods for minor species. Notably, SZ shows a noticeable error in predicting mean and variance of QoI of the minor species. In contrast, the result with GBATC reasonably tracks the overall variations in a qualitative manner. The main reason why minor species is likely to exhibit a larger error is because their concentration is relatively-low compared to the major species, and as such, a small error in PD can significantly affect the computation of QoI. Moreover, the spatio-temporal evolution of minor species are likely to be more abrupt compared to that of major species. This makes the spatio-temporal correlation used in this study less effective for minor species. Given that an accurate reproduction of the mean and variance profiles of the species is essential for understanding turbulence-chemistry interactions in turbulent flames, this result illustrates that GBATC is a robust tool for domain scientists to utilize in their large-scale and multiphysics simulations.

In future work, the efficiency of GBATC will be investigated over the entire time span of the DNS dataset, $t$ = 0--3 ms, particularly when low-temperature and high-temperature ignition of the lean $n-$heptane+air mixture is observed (low-temperature and high-temperature ignitions are observed near at 0.2--0.4 ms and 2.3--2.5 ms, respectively). 

\section{Related Work}\label{sec:relwork}
The literature on image and video compression techniques is large: please see survey articles: \cite{saha2000image} for image compression; \cite{bhaskaran1997image} for video compression; and \cite{ma2019image} for more recent neural network-based compression. We mainly focus on reduction and compression approaches for scientific applications because techniques for image and video compression are not directly applicable as discussed earlier.

Error-bounded lossy compression is considered as the most effective compression technique because it provides reliability that is useful for scientific applications.  ZFP \cite{fox2020stability} is a transform-based compression model that splits a dataset into a set of unoverlapped $4^{d}$ blocks, where $d$ is the data dimensionality. Each block is decorrelated using a nearly orthogonal transform.  TTHRESH \cite{ballester2019tthresh} is a dimensionality reduction-based model and uses higher-order singular value decomposition (HOSVD) to reduce the dimension of the data according to importance. FAZ \cite{liu2023faz} is a comprehensive compression framework that has functional modules and leverages prediction models and wavelets. MGARD \cite{MGARD_2,MGARD_3} provides error-controlled lossy compression using a multigrid approach. It transforms floating-point scientific data into a set of multilevel coefficients. Several of the above methods use quantization and encoding to further reduce the amount of storage required while adhering to the user's specified error bounds on PD.  

In \cite{Archibald2023}, the weak SINDy algorithm
\cite{WeakSINDy1,WeakSINDy2} that estimates and identifies an underlying
dynamic system and an orthogonal decomposition are combined to compress
streaming scientific data. 
To compress data generated from partial differential equation (PDE)
simulations, QuadConv \cite{QuadConv} is developed based on convolutional
autoencoders that perform convolution via quadrature for non-uniform and mesh-based
data. SPERR \cite{SPERR} is  a transform-based compression algorithm that uses a multilevel discrete
wavelet transform to decorrelate the data. The work in \cite{Lindstrom1,Lindstrom2}
utilizes progressive data compression  for adaptive handling
of compressed data according to a post-processing task.   SRN-SZ \cite{liu2023srn} is noteworthy in its marriage of ML based super-resolution with SZ. 

Recently, several lossy compressors have advanced the error control onto downstream QoI. MGARD derives a norm based on the multigrid theory, which can be used to ensure the error preservation for linear QoI \cite{MGARD_3}, and newer techniques support QoIs that are a combination of polynomials and radicals  \cite{xuan2024}. A variation of SZ has also been proposed \cite{jiao2022toward}, which relies on a pre-evaluation of target QoIs and deriving pointwise PD error bounds that will provide guarantees on QoIs. However, this requires knowing the original QoI values and is only applicable to univariate QoIs. Several additional compression methods have been developed to reduce the data while preserving topological features such as critical points \cite{liang2022toward}, but they do not generalize to other QoIs. Although, our GAE can be easily augmented to provide guarantees on PD \cite{lee2023nonlinear}, linear QoIs \cite{lee2023nonlinear}, and nonlinear QoIs\cite{lee2023nonlinear} (where the error on the latter can be close to floating point errors).
However, none of these methods directly address $O(N)$ QoI that are represented by the reaction rates for each tensor of size $N$.

This is part of our future endeavors and not supported by any prior work.

\section{Conclusions}\label{sec:conclusion}
The proposed guaranteed block autoencoder (GBATC)  utilizes a multidimensional block of tensors in space and time to capture the spatiotemporal relationships as well as the interrelationships within the tensor that are present in many CFD applications. 
A convolutional autoencoder is then utilized to capture the spatiotemporal correlations within each block.
 To improve the compression quality further, we introduce a tensor correction network. After training the AE, we
convert each reconstructed instance by the AE, comprising S
number of 3D blocks, into a set of S-dimensional tensors.
The tensor correction network takes the reconstructed tensors and learns a reverse point-wise (in temporal and spatial space) mapping from the reconstructed tensors to the original tensors.
Our approach also provides guarantees on PD \cite{lee2023nonlinear}. To guarantee the error bound of the reconstructed data, principal component analysis (PCA) is applied to the residual between the original and reconstructed data. This yields a basis matrix, which is then used to project the residual at each instance. The resulting coefficients are retained to enable accurate recovery of the residual. The number of coefficients saved is incrementally increased until the error bound is satisfied. Additionally, quantization and entropy coding techniques are applied to both the latent data from the GAE and the PCA coefficients. This further improves the compression ratio of the overall process.

We validate our method using the simulation data generated by Sandia's compressible reacting direct numerical simulation (DNS) code, S3D \cite{Chen09}.  Experimental results demonstrate the our approach can derive two orders of magnitude in reduction while still maintaining the errors of primary data under scientifically acceptable bounds.
In comparison to previous research \cite{SZ_3}, our method achieves a substantially higher compression ratio.  

It is also important that the reduction methods provide reasonable errors on downstream QoIs.  One of the crucial QoIs for the species transport equation is the net production rate for each species (and this involves other species as well), with the rate being dependent on the forward and reverse rate constants of the reactions underlying the production.  We empirically show that our approach performs considerably better than SZ on these QoIs. Although our previous work can guarantee linear QoIs \cite{lee2023nonlinear}, and nonlinear QoIs \cite{lee2023nonlinear} (where the error on the latter can be close to floating point errors), they are applicable to only $O(1)$ QoIs for tensors of size $N$ and cannot be easily extended to $O(N)$ QoIs. Also, we separate out the AE and tensor correction network in terms of training. The additional network can be incorporated into the AE decoder and can be trained within the AE using an end-to-end approach. This will be part of our future work.

\section*{Acknowledgment}
This work was partially supported by DOE RAPIDS2 DE-SC0021320 and DOE DE-SC0022265.  The work contributed by Sandia was supported by the DOE Office of Science Distinguished Scientists Fellow Award. Sandia National Laboratories is a multimission laboratory managed and operated by National Technology and Engineering Solutions of Sandia, LLC., a wholly owned subsidiary of Honeywell International, Inc., for the U.S. Department of Energy’s National Nuclear Security Administration under contract DE-NA-0003525.


\bibliographystyle{IEEEtran}
\bibliography{ref,ref2,chen,references2,DOECompressionML,klaskybib,UFL1,gong-bib,Jong}

\end{document}